\documentclass{article}
\usepackage{ijcai24}

\usepackage{times}
\usepackage{soul}
\usepackage{url}
\usepackage[hidelinks]{hyperref}
\usepackage[utf8]{inputenc}
\usepackage[small]{caption}
\usepackage{graphicx}
\usepackage{booktabs}
\usepackage{algorithm}
\usepackage{algorithmic}
\usepackage[switch]{lineno}

\usepackage{amsfonts}       
\usepackage{amssymb}
\usepackage{amsmath}
\usepackage{amsthm}
\usepackage{mathtools}
\usepackage{latexsym,bm}
\DeclareMathOperator{\E}{\mathbb{E}}
\DeclareMathOperator{\R}{\mathbb{R}}
\usepackage{multicol}
\usepackage{multirow}
\usepackage[table,xcdraw]{xcolor}

\usepackage{color}

\newtheorem{theorem}{Theorem}
\theoremstyle{remark}
\newtheorem{remark}[theorem]{Remark}

\usepackage{scrextend}

\title{SpecSTG: A Fast Spectral Diffusion Framework \\
for Probabilistic Spatio-Temporal Traffic Forecasting}

\author{
Lequan Lin$^1$\thanks{The corresponding author.}\footnote{Equal contributions.}
\and
Dai Shi$^1$\footnotemark[\value{footnote}]
\and
Andi Han$^2$
\And
Junbin Gao$^1$\\
$^1$The University of Sydney\\
$^2$Riken AIP\\
\emails
\{lequan.lin, dai.shi, junbin.gao\}@sydney.edu.au,
andi.han@riken.jp
}

\begin{document}

\maketitle

\begin{abstract}
Traffic forecasting, a crucial application of spatio-temporal graph (STG) learning, has traditionally relied on deterministic models for accurate point estimations. Yet, these models fall short of quantifying future uncertainties. Recently, many probabilistic methods, especially variants of diffusion models, have been proposed to fill this gap. However, existing diffusion methods typically deal with individual sensors separately when generating future time series, resulting in limited usage of spatial information in the probabilistic learning process. In this work, we propose SpecSTG, a novel spectral diffusion framework, to better leverage spatial dependencies and systematic patterns inherent in traffic data. More specifically, our method generates the Fourier representation of future time series, transforming the learning process into the spectral domain enriched with spatial information. Additionally, our approach incorporates a fast spectral graph convolution designed for Fourier input, alleviating the computational burden associated with existing models. Compared with state-of-the-arts, SpecSTG achieves up to 8\% improvements on point estimations and up to 0.78\% improvements on quantifying future uncertainties. Furthermore, SpecSTG's training and validation speed is 3.33$\times$ of the most efficient existing diffusion method for STG forecasting. The source code for SpecSTG is available at \textcolor{blue}{\url{https://anonymous.4open.science/r/SpecSTG}}.
\end{abstract}

\section{INTRODUCTION}

\begin{figure}[t]
\begin{minipage}[b]{1.0\linewidth}
  \centering
  \centerline{\includegraphics[height = 4cm, width=8.5cm]{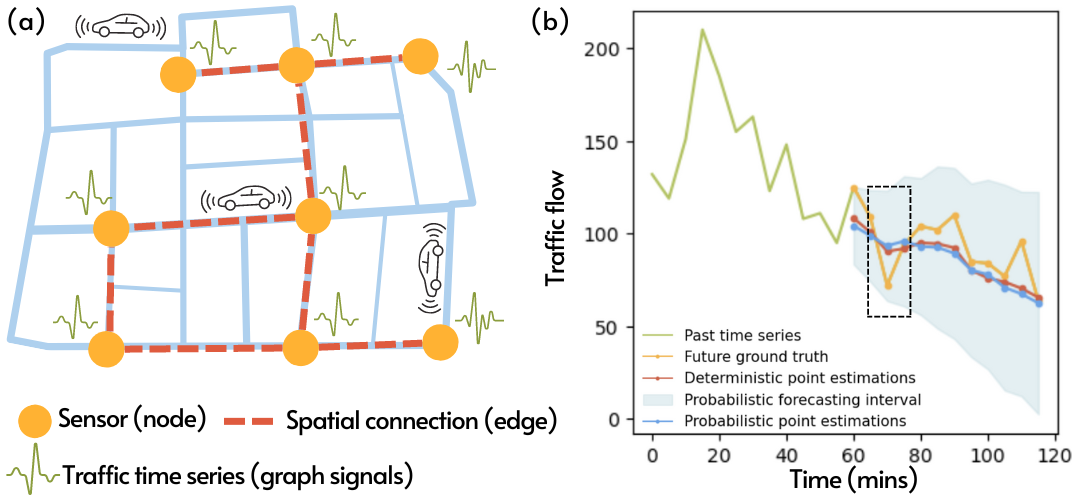}}
\end{minipage}
\caption{Illustrations: (a) an example of traffic STG; (b) traffic flow forecasting in future 60 minutes with GMAN (deterministic) and SpecSTG (probabilistic) on PEMS04.}
\label{fig:stgexample}
\end{figure}

\begin{figure*}[t!]
\begin{minipage}[b]{1.0\linewidth}
  \centering
  \centerline{\includegraphics[height = 5cm, width=17.7cm]{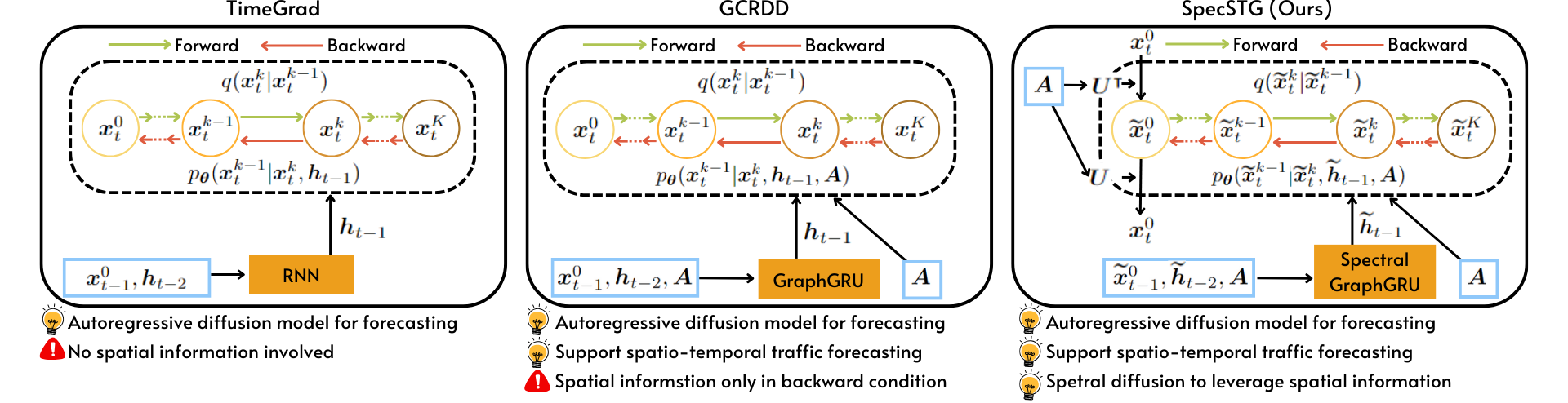}}
\end{minipage}
\caption{The overview of SpecSTG. Illustrations of TimeGrad and GCRDD are provided to show the novelty and advantage of our approach.}
\label{fig:specstgoverview}
\end{figure*}

Traffic forecasting on road networks is an essential application domain of spatio-temporal graph (STG) learning~\cite{yuan2021survey,jin2023spatio,jin2023survey}. As a crucial component of the Intelligence Transportation System (ITS), accurate and informative prediction of future traffic dynamics provides essential guidelines to traffic authorities in decision-making~\cite{lana2018road,boukerche2020artificial}. Since traffic data, such as vehicle speed and traffic flow, are collected from sensors in the continuous space of road networks, they present strong spatio-temporal dependencies, especially at neighbouring locations and time windows~\cite{guo2019attention}. This naturally leads to their representation as STGs: the traffic network is modelled as a graph, in which nodes are sensors and edges are decided with some criteria such as geographic distances. Hence, temporal records are stored as graph signals, while spatial information is encapsulated in the graph structure. In \textbf{Figure \ref{fig:stgexample} (a)}, we provide a visualized example of traffic STG.

STG traffic forecasting aims at predicting future values at all sensors based on past time series and spatial connections in the traffic network. This task has traditionally relied on deterministic models such as DCRNN~\cite{li2017diffusion} and GMAN~\cite{zheng2020gman} to produce accurate point estimations. Nevertheless, these models may fall short in identifying unexpected variations that lead to consequential change in traffic regulations~\cite{pal2021rnn,wen2023diffstg,hu2023towards}. This limitation can be overcome by probabilistic methods, which alternatively approximate the distribution of future time series, thus leading to more uncertainty-aware predictions. For example, \textbf{Figure \ref{fig:stgexample} (b)} shows the results of deterministic and probabilistic models on the PEMS04 traffic flow forecasting task~\cite{guo2019attention}. The deterministic model can only provide point estimations (red), while the probabilistic model is capable of generating both point estimations and the forecasting interval (blue) which captures some abrupt fluctuations in traffic flow (black box).

Among all the probabilistic models applicable to STG traffic forecasting, we specifically focus on diffusion models~\cite{yang2023diffusion,ho2020denoising,lin2023diffusion}. Classic diffusion models for time series forecasting, such as TimeGrad \cite{rasul2021autoregressive}, generally follow a forward-backward training process: the generative target (i.e., future time series) is first turned into white noise and then recovered 
by a learnable backward kernel. The backward kernel is a conditional distribution which is similar to traditional diffusion models except for the inclusion of encoded past temporal information in its conditions. 
For prediction, samples from the distribution of future time series are generated by denoising white noise with the learned backward kernel. Since such methods are incapable of handling spatial information in STGs, diffusion models for STG forecasting, such as DiffSTG~\cite{wen2023diffstg} and GCRDD~\cite{ruikun2023}, incorporate graph structure in the backward kernel condition. More specifically, the backward kernel is usually approximated by a denoising network, which takes graph structure and other conditions as input to predict noises injected in the forward process.

Our work considers two limitations of existing diffusion methods for STG forecasting.
(\textbf{Limitation \#1}) Although these methods emphasize on the importance of spatial information in STG forecasting, they only use  spatial information in the backward kernel condition. 
Consequently, the involvement of spatial information in the overall forward-backward diffusion learning process is limited. 
(\textbf{Limitation \#2}) The denoising network of existing methods usually relies substantially on graph convolution to encode spatial information. This often introduces a complexity of $\mathcal{O}(N^2)$, quadratic in the number of sensors $N$ in a traffic STG (see Subsection \ref{sec:spectral_graph_conv} for more details).
Thus the computational cost is high especially for large traffic networks, leading to slow training and sampling.

In this work, we propose a novel spectral diffusion framework (SpecSTG), which adopts the graph Fourier representation of time series as the generative target. According to the spectral graph theory, the graph Fourier representation is a measurement of variations in graph signals guided by the graph structure \cite{chung1997spectral,kreuzer2021rethinking}. In the context of STGs, we may treat time points as features, thereby the graph Fourier representation can be considered as a new time series of systematic fluctuations enriched by spatial information.   
Hence, \textbf{Limitation \#1} is resolved by generating the Fourier representation rather than the original time series, transforming the entire diffusion process into the spectral domain. This effectively leverages the graph structure to construct a more comprehensive diffusion base with additional systematic and spatial patterns. Besides, with no loss of information, the generated data can be converted back to the original domain via the inverse Fourier transform for prediction. 
\textbf{Limitation \#2} is mitigated by replacing the graph convolution with a light-complexity alternative, which only works for the Fourier input (details in Subsection \ref{sec:spectral_graph_conv}). 
An overview of SpecSTG can be found in \textbf{Figure \ref{fig:specstgoverview}}. We also provide illustrations of TimeGrad and GCRDD for comparison.
The contributions of this paper are three-fold: 
\begin{enumerate}
    \item To our best knowledge, this is the first work that explores probabilistic STG forecasting on the graph spectral domain;
    \item SpecSTG achieves up to 8\% improvements on point estimations and up to 0.78\% improvements on generating compatible forecasting intervals. 
    \item SpecSTG's training and validation speed is 3.33$\times$ of the most efficient existing diffusion method for STG forecasting. Additionally, SpecSTG significantly accelerates the sampling process, particularly for large sample sizes.
\end{enumerate}

\section{\textsc{PRELIMINARIES}}
\subsection{Spatio-Temporal Graphs}
STGs can be considered as a multidimensional graph representation of entities in a system with time series as graph signals. In traffic forecasting, we model sensors as nodes and then create edges based on some spatial relationships such as geographic distances. The average traffic records in observation periods are modelled as graph signals. 
For a traffic network with $N$ sensors, the corresponding STG can be denoted as $\mathcal{G}\{\mathcal{V},\mathcal{E},\bm{A}\}$, where $\mathcal{V}$ is the set of nodes/sensors, $\mathcal{E}$ is the set of edges, and $\bm{A} \in \mathbb{R}^{N\times N}$ is the adjacency matrix. \textcolor{black}{$\bm{A}$ is assumed to be undirected and can be either weighted or unweighted.} The graph signals are denoted as $\bm{X}_{\mathcal{G}} = \{\bm{x}_1, \bm{x}_2, ..., \bm{x}_t, ... | \bm{x}_t \in \R^{N \times D_x} \}$, where $D_x$ is the number of variables. In traffic forecasting, it is common that only one variable such as speed or flow is of interest~\cite{li2017diffusion,guo2019attention}, thus we have $D_x = 1$ such that $\bm{x}_t \in \R^{N}$. 

\subsection{Spatio-Temporal Graph Forecasting}

The objective of STG traffic forecasting is to predict a future time series window $\bm{X}_f= \{\bm{x}_{t_0+1}, \bm{x}_{t_0+2}, ..., \bm{x}_{t_0+f}\}$ given the past context window $\bm{X}_c = \{\bm{x}_{t_0-c+1}, \bm{x}_{t_0-c+2}, ..., \bm{x}_{t_0}\}$, where $f$ and $c$ are the length of future and past windows.  
We denote the combination of past and future time series as $\bm{X} = \{\bm{x}_{t_0-c+1}, \bm{x}_{t_0-c+2},..., \bm{x}_{t_0+f}\}$. 
Normally, the target distribution of generative models depends on the sampling methods: one-shot methods produce all future predictions together from the distribution $q(\bm{X}_f|\bm{X}_c, \bm{A})$~\cite{wen2023diffstg,liu2023pristi}, while autoregressive methods generate samples from $q(\bm{x}_{t}|\bm{X}_{t_0-c+1:t-1}, \bm{A})$ for $t = t_0+1, t_0+2, ..., t_0+f$ successively, where $\bm{X}_{t_0-c+1:t-1} = \{\bm{x}_{t_0-c+1}, ..., \bm{x}_{t-1}\}$~\cite{ruikun2023}. Autoregressive methods often capture the sequential information in consecutive time points more closely. However, they are associated with higher time costs because of the non-parallel step-by-step sampling process. Our method, SpecSTG, is formulated within the autoregressive framework but is equipped with a specially designed spectral graph convolution to mitigate computational inefficiency.

\subsection{Denoising Diffusion Probabilistic Model}
Denoising diffusion probabilistic models (DDPMs)  learn how to generate samples from the target distribution via a pair of forward-backward Markov chains \cite{ho2020denoising,yang2023diffusion}. Assuming that $\bm{x}^0 \sim q(\bm{x}^0)$ is the original data, for diffusion step $k = 0,1, ..., K$, the forward chain injects Gaussian noises to $\bm{x}^k$ until $q(\bm{x}^K) \coloneqq \int q(\bm{x}^K|\bm{x}^0) q(\bm{x}^0) \mathrm{d}\bm{x}^0 \approx \mathcal{N}(\bm{x}^K; \bm{0}, \bm{I})$. As a special property, given a noise schedule $\beta = \{\beta_1, \beta_2, ...,\beta_K\}$, we may directly compute the disturbed data at step $k$ as $\bm{x}^k = \sqrt{\Tilde{\alpha}_k} \bm{x}^0 + \sqrt{1-\Tilde{\alpha}_k} \bm{\epsilon}$, where $\Tilde{\alpha}_k = \prod_{i=1}^k (1 - \beta_i)$ and  $\bm{\epsilon} \sim \mathcal{N} (\bm{0},\bm{I})$. Next, the backward chain denoises from $\bm{x}^K$ to recover $p_{\bm{\theta}}(\bm{x}^0)$ through a probabilistic backward kernel $p_{\bm{\theta}}(\bm{x}^{k-1}|\bm{x}^{k})$, where $\bm{\theta}$ denotes all learnable parameters. In practice, the backward kernel is usually optimized with a denoising network $\bm{\epsilon}_{\bm{\theta}}$ by minimizing the loss function
\begin{equation}\label{eq_DDPMloss2}
    \mathcal{L}_{DDPM}(\bm{\theta}) = \E_{k, \bm{x}^0, \bm{\epsilon}}  \left\|\bm{\epsilon} - \bm{\epsilon}_{\bm{\theta}} \left(\bm{x}^k, k \right)\right\|^2,
\end{equation}
where $\bm{\epsilon} \sim \mathcal{N} (\bm{0},\bm{I})$ represents the noises injected in the forward diffusion steps. Please refer to \textbf{Appendix A} for more details on DDPMs.

\section{\textsc{SPATIO-TEMPORAL GRAPH FOURIER TRANSFORM}}
Given a traffic STG $\mathcal{G}\{\mathcal{V}, \mathcal{E}, \bm{A}\}$ with $N$ sensors, the normalized graph Laplacian is computed as $\bm{L} = \bm{I}_N - \bm{D}^{-\frac{1}{2}}\bm{A}\bm{D}^{-\frac{1}{2}}$, where $\bm{I}_N \in \R^{N \times N}$ is the identity matrix, and $\bm{D} \in \R^{N \times N}$ is the diagonal degree matrix.
We denote the eigen-decomposition of the graph Laplacian as $\bm{L} = \bm{U} \bm{\Lambda} \bm{U}^\intercal$, where $\bm{U}\in \R^{N \times N}$ and $\bm{\Lambda}\in \R^{N \times N}$ are the corresponding eigenvector and eigenvalue matrices, respectively.
Considering the univariate graph signal $\bm{X}_{\mathcal{G}} = \{\bm{x}_1, \bm{x}_2, ..., \bm{x}_t, ...| \bm{x}_t \in \R^N\}$, the Fourier transform for each time point $t$ is given by $\Tilde{\bm{x}}_t = \bm{U}^\intercal \bm{x}_t$,
known as the Fourier representation of $\bm{x}_t$ in the spectral domain. 
The Fourier reconstruction is $\bm{x}_t = \bm{U} \Tilde{\bm{x}}_t$. 
The orthonormal $\bm{U}$ ensures a lossless reconstruction for the temporal information. We may compute in matrix form for all time points as $\Tilde{\bm{X}} = \bm{U}^\intercal \bm{X}$ and $\bm{X} = \bm{U} \Tilde{\bm{X}}$. In \textbf{Appendix B}, we also discuss how to naturally extend the method to multivariate traffic STGs with $\bm{x}_t \in \R^{N \times D_x}$, $D_x \geq 2$.

The graph Fourier transform can be understood as a projection of graph signals onto the spectral domain spanned by the eigenvector basis of graph Laplacian. The operator $\bm{U}$ brings rich positional information for graph signals and offers a platform to investigate the variations among signals through a global perspective on the graph. In particular, when the input is a traffic STG, the Fourier representation measures how graph signals (time series values) fluctuate across the network. This effectively integrates spatial connectivity into time series, leading to a spatial-aware forecasting paradigm.

\section{\textsc{THE PROPOSED METHOD}}\label{sec:method}
SpecSTG assumes that the Fourier representation of future time series follows the distribution
\begin{equation} q(\Tilde{\bm{X}}_f^0|\Tilde{\bm{X}}_c^0,  \bm{A}) \approx    \prod_{t = t_0+1}^{t_0+f} p_{\bm{\theta}}(\Tilde{\bm{x}}_{t}^0|\Tilde{\bm{h}}_{t-1}, \bm{A}),
\end{equation}
where $\Tilde{\bm{X}}_c^0$ 
and $\Tilde{\bm{X}}_f^0$ are the noise-free Fourier representations of past and future time series, respectively. $\Tilde{\bm{h}}_{t-1}$ represents past spatio-temporal condition encoded by a spectral recurrent encoder.  
For each time point $t = t_0+1,...,t_0+f$, the diffusion process will learn the corresponding backward kernel $p_{\bm{\theta}}(\Tilde{\bm{x}}_t^{k-1}|\Tilde{\bm{x}}_t^k,\Tilde{\bm{h}}_{t-1},\bm{A})$ with $k = 1, ..., K$. The objective function of SpecSTG is given by
\begin{equation}\label{objective}
    \mathcal{L}(\bm{\theta}) = \E_{t, k, \Tilde{\bm{x}}_t^0,  \bm{\epsilon}_t} \left\|\bm{\epsilon}_t - \bm{\epsilon}_{\bm{\theta}} \left(\Tilde{\bm{x}}_t^k, k, \Tilde{\bm{h}}_{t-1}, \bm{A} \right)\right\|^2,
\end{equation}
where $t$ denotes future time points, $k$ denotes diffusion steps, $\Tilde{\bm{x}}_t^k$ is the disturbed data at time point $t$ and step $k$, and $\bm{\epsilon}_t \sim \mathcal{N}(\bm{0}, \bm{I})$. It is worth noting that $t$ may also start with $t_0-c+1$ instead of $t_0+1$ to facilitate the learning of autoregressive dependencies by considering both past and future instances. Lastly, we employ a graph-modified WaveNet architecture~\cite{oord2016wavenet} for the denoising network $\bm{\epsilon}_{\bm{\theta}}(\cdot)$, which is specially designed for Fourier input. In the rest of this section, we will focus on details of SpecSTG components, training, and inference. The visualization of SpecSTG architecture can be found in \textbf{Figure \ref{fig:specstgoverview}}.

\subsection{Light-Complexity Spectral Graph Convolution}\label{sec:spectral_graph_conv}

\noindent Diffusion models for STG forecasting usually rely on graph convolution to encode spatial information in the condition of backward kernel~\cite{ruikun2023,wen2023diffstg,liu2023pristi}. Straightforwardly, we adopt the spectral convolution, which typically transforms graph signals to the spectral domain and then processes the data via frequency filtering before they are converted back to the original domain~\cite{defferrard2016convolutional,kipf2016semisupervised}. However, with SpecSTG, the Fourier representation is already formed as input, so the transform is no longer required in the convolution. In addition, to ensure that the model pipeline flows in the spectral domain throughout the diffusion learning process, we do not apply reconstruction either. 

We choose the Chebyshev convolution~\cite{defferrard2016convolutional}, whose filters are formed with Chebyshev polynomials to accelerate computation as $\mathrm{ChebConv}(\bm{X}) = \bm{U} \sum_{j = 0}^{J-1} \bm{\phi}_j \mathcal{T}_j(\Tilde{\bm{\Lambda}}) \bm{U}^\intercal\bm{X}$,
for polynomial degrees up to $J-1$, where $\mathcal{T}_j(\Tilde{\bm{\Lambda}}) \in \mathbb{R}^{N\times N}$ is the $j$-th order Chebyshev polynomial evaluated at $\Tilde{\bm{\Lambda}} = 2\bm{\Lambda}/\lambda_{max} - \bm{I}_N$ with $\bm{\phi}_j$ being a learnable coefficient. With Fourier representation $\Tilde{\bm{X}} = \bm{U}^\intercal\bm{X}$ as input, we define the modified spectral graph convolution as
\begin{equation}\label{specconv}
    \mathrm{SpecConv}(\Tilde{\bm{X}}) \coloneqq \sum_{j = 0}^{J-1} \bm{\phi}_j \mathcal{T}_j(\Tilde{\bm{\Lambda}}) \Tilde{\bm{X}}.
\end{equation}
\begin{remark}[Complexity Analysis]
 Our spectral convolution has only $\mathcal{O}(N)$ complexity, contrasting to the $\mathcal{O}(N^2)$ complexity of the classic Chebyshev convolution. This significant improvement endows SpecSTG with the advantage of time efficiency over methods using Chebyshev convolution, such as GCRDD. Other methods such as DiffSTG may use the graph convolution in~\cite{kipf2016semisupervised} as $\mathrm{GCNConv}(\bm{X}) = \bm{A}\bm{X}$, whose complexity is $\mathcal{O}(
|\mathcal{E}|)$ where $|\mathcal{E}|$ is the number of edges.  As we see in Section \ref{dataset}, $|\mathcal{E}|$ is often much larger than $N$ in traffic STGs.   
\end{remark}


\subsection{Spectral Recurrent Encoder}\label{sec:SG-GRU}

We design $\mathrm{SG\text{-}GRU}$ as the spectral version of Graph GRU \cite{seo2018structured} to encode past time series and spatial information in graph Fourier domain as follows. For $t = t_0 - c +1, ..., t_0$:
\begin{align}
    &\bm{z} = \sigma(\mathrm{SpecConv}(\Tilde{\bm{x}}^0_t)\bm{W}_{z_1} \!+ \mathrm{SpecConv}(\Tilde{\bm{h}}_{t-1})\bm{W}_{z_2})\\
    &\bm{r} = \sigma (\mathrm{SpecConv}(\Tilde{\bm{x}}^0_t)\bm{W}_{r_1} \!+ \mathrm{SpecConv}(\Tilde{\bm{h}}_{t-1})\bm{W}_{r_2})\\
    &\bm{\zeta} = \mathrm{tanh}(\mathrm{SpecConv}(\Tilde{\bm{x}}^0_t)\bm{W}_{\zeta_1} \!+ \mathrm{SpecConv}(\bm{r}\odot\Tilde{\bm{h}}_{t})\bm{W}_{\zeta_2})\\
    &\Tilde{\bm{h}}_{t+1}  = \bm{z} \odot \Tilde{\bm{h}}_{t} \!+ (1 - \bm{z}) \odot \bm{\zeta},
\end{align}
where $\Tilde{\bm{h}}_0 \in \mathbb{R}^{D_h}$ is a vector of zeors with $D_h$ being the hidden size, $\bm{W}_{z_1},  \bm{W}_{r_1}, \bm{W}_{\zeta_1} \in \R^{1 \times D_h},$ $\bm{W}_{z_2},  \bm{W}_{r_2}, \bm{W}_{\zeta_2} \in \R^{D_h \times D_h}$ are learnable weights included in $\bm{\theta}$, and $\sigma$ is the sigmoid activation function. $\bm{z}$ and $\bm{r}$ are known as update gate and reset gate. $\bm{\zeta}$ is the candidate state storing current information to be updated in the hidden state. In the implementation, we may also input time features $\bm{\Gamma} = \{\bm{\gamma}_{t_0-c+1}, \bm{\gamma}_{t_0-c+2}, ..., \bm{\gamma}_{t_0+f}\}$ such as day of week and week of month by concatenating them with $\Tilde{\bm{X}}$.

\subsection{Denoising Network: Spectral Graph WaveNet}

\begin{figure}[t]
\begin{minipage}[b]{1.0\linewidth}
  \centering
  \centerline{\includegraphics[height = 5.5cm, width=6.5cm]{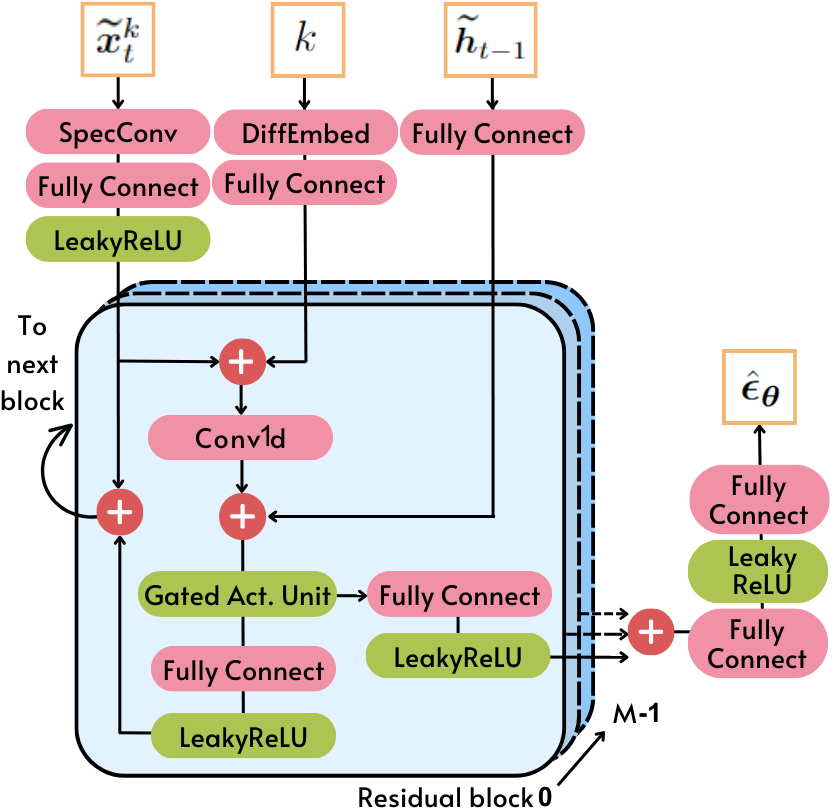}}
\end{minipage}
\caption{The denoising network $\bm{\epsilon}_{\bm{\theta}}$ is a modified WaveNet structure.}
\label{fig:sgwave}
\end{figure}

In reminiscent of TimeGrad, we design \underline{\textbf{S}}pectral \underline{\textbf{G}}raph \underline{\textbf{Wave}}Net ($\text{SG-Wave}$) as the $\bm{\epsilon}_{\bm{\theta}}$ of SpecSTG (see details in \textbf{Figure \ref{fig:sgwave}}). Besides, we replace some $\mathrm{Conv1d}$ layers with fully connected linear layers for efficient training and sampling. The network takes disturbed data $\Tilde{\bm{x}}_t^k$, diffusion step $k$, hidden condition $\Tilde{\bm{h}}_{t-1}$, and graph adjacency $\bm{A}$ as inputs, and aims at predicting the noise $\bm{\epsilon}_t \sim \mathcal{N}(\bm{0},\bm{I})$ at step $k$ for time point $t$. 


\begin{algorithm}[t]
\caption{SpecSTG training}
\label{alg1} 
\textbf{Input:}
The distribution of training data after Fourier transform: $q(\{\Tilde{\bm{X}}_c, \Tilde{\bm{X}}_f\})$; hidden states: $\Tilde{\bm{h}}$; number of diffusion steps: $K$; noise schedule $\{\beta_1, \beta_2, ..., \beta_K$\}, graph: $\mathcal{G}(\mathcal{V}, \mathcal{E},\bm{A})$.
\begin{flushleft}
\textbf{Output:} Optimized denoising network $\bm{\epsilon}_{\bm{\theta}}$.
\end{flushleft}
\begin{algorithmic}[1]
\STATE Sample $\{\Tilde{\bm{X}}_c, \Tilde{\bm{X}}_f\} \sim q(\{\Tilde{\bm{X}}_c, \Tilde{\bm{X}}_f\})$
\WHILE{$\textit{Not Convergence}$}
\FOR{$t = t_0+1, t_0+2, ..., t_0+f$}
\STATE $k \sim \mathrm{Uniform}(1,K)$, $\bm{\epsilon}_t \sim \mathcal{N}(\bm{0}, \bm{I})$
\STATE Compute $\Tilde{\bm{x}}_{t}^k = \sqrt{\Tilde{\alpha}_k} \Tilde{\bm{x}}_t^0 + \sqrt{1-\Tilde{\alpha}_k} \bm{\epsilon}_t$
\ENDFOR
\STATE Take gradient step on and do gradient descent for $\theta$
\begin{equation*}
    \nabla_{\bm{\theta}} \E_{t, k, \Tilde{ \bm{x}}_t^0, \bm{\epsilon}_t} \left\|\bm{\epsilon}_t - \bm{\epsilon}_{\bm{\theta}} \left(\Tilde{\bm{x}}_t^k, k, \Tilde{\bm{h}}_{t-1}, \bm{A} \right)\right\|^2
\end{equation*} 
\ENDWHILE
\end{algorithmic}
\end{algorithm}

\begin{algorithm}[t]
\caption{SpecSTG sampling and prediction for $\bm{x}_{t}^0$}
\label{alg2} 
\textbf{Input:} Hidden state: $\Tilde{\bm{h}}_{t-1}$; Fourier operator $\bm{U}$; number of samples: $S$; variance hyperparameter: $\sigma_k$.

\textbf{Output:} Prediction $\hat{\bm{x}}_{t}^0$.
 
\begin{algorithmic}[1]
\STATE
Randomly generate $S$ samples $\{\Tilde{\bm{x}}_{t,s}^K\}^S_{s=1} \sim \mathcal{N}(\bm{0},\bm{I})$ and do the follows in parallel for all $s$.

\FOR{$k = K, K-1, ..., 1$}
\STATE {$\bm{e} = \bm{0}$ if $k = 1$ else $\bm{e} \sim \mathcal{N}(\bm{0},\bm{I})$} 

\STATE Compute and update $\Tilde{\bm{x}}_{t,s}^{k-1}$ as
\begin{flalign*}
\hspace{-1.3cm}\quad\quad\quad\frac{1}{\sqrt{1\!-\beta_{k}}} 
    \big(
    \Tilde{\bm{x}}_{t,s}^{k}\!-\!\frac{
    \beta_{k}}{\sqrt{1\!-\Tilde{\alpha}_{k}}}
    \bm{\epsilon}_{\bm{\theta}}(\Tilde{\bm{x}}_{t,s}^{k},k,\Tilde{\bm{h}}_{t-1},\bm{A})\big)+\sigma_k \bm{e}
\end{flalign*}
\ENDFOR
\STATE Take average on $S$ samples $\Tilde{\bm{x}}_{t}^0 = \frac{1}{S} \sum_{s =1}^S \Tilde{\bm{x}}_{t,s}^0$
\STATE Compute predictions $\hat{\bm{x}}^0_t = \bm{U}\Tilde{\bm{x}}_{t}^0$
\end{algorithmic}
\end{algorithm}

\subsection{Training and Inference}
We compute the Fourier representation of all training data before the training process.
Next, we input randomly sampled $\{\Tilde{\bm{X}}_c, \Tilde{\bm{X}}_f\}$ to $\text{SG-GRU}_{\bm{\theta}}$ to obtain hidden states $\Tilde{\bm{h}} = \{\Tilde{\bm{h}}_{t_0}, ..., \Tilde{\bm{h}}_{t_0+f-1}\}$. 
During training, with a pre-specified noise schedule $\{\beta_1, \beta_2, ..., \beta_K\}$, we randomly sample noise $\bm{\epsilon}_t \sim \mathcal{N}(\bm{0},\bm{I})$ and step $k \sim \mathrm{Uniform}(0,K)$ to compute disturbed data $\Tilde{\bm{x}}_t^k$ for $t = t_0+1, t_0+2, ..., t_0+f$ . Finally, we take a gradient step on the objective function in Equation (\ref{objective}). The training algorithm is provided in \textbf{Algorithm \ref{alg1}}. With the denoising network, we can generate samples and make predictions for the forecasting task. The generation process adopts autoregressive sampling, which means we generate samples for each time point one by one. For example, after we generate samples for $t=t_0+1$, we feed the sample mean back to the SG-GRU module to compute $\Tilde{\bm{h}}_{t_0+1}$, and then use it to generate samples for $t = t_0+2$. Lastly, we convert the predictions back to the original domain via Fourier reconstruction. The sampling and prediction algorithm for a one-time point $\bm{x}_t^0$ is presented in \textbf{Algorithm \ref{alg2}}.

\section{\textsc{EXPERIMENTS}}
\subsection{Datasets and Baseline Models}\label{dataset}
We validate our model on two traffic datasets, PEMS04 and PEMS08 \cite{guo2019attention}, collected by the California Transportation Agencies' (CalTrans) Performance Measurement System (PEMS)~\cite{chen2001freeway}. PEMS04 comprises traffic records from 307 sensors in California's District 04 from Jan 1st to Feb 28th, 2018, while PEMS08 includes data from 170 sensors in District 08 from July 1st to 31st August 2018. Our experiments focus on traffic flow and speed available in both datasets, denoted as ``F'' and ``S'' respectively. For instance, ``PEMS04F'' denotes traffic flow records in the PEMS04 dataset. Each time point covers 5 minutes, with the corresponding value representing the average records during that interval. The speed data are continuous, allowing us to introduce random Gaussian noises. Although traffic flow (i.e., the number of vehicles) is a discrete variable, we still treat it as continuous considering the fact that it contains numerous unique values. More details about the datasets can be found in \textbf{Table \ref{tab:datasets}}.

\begin{table}[h] 
    \centering
    \caption{Dataset details.}
    \begin{tabular}{llrrr}
        \toprule
        Dataset & Type & \#Nodes & \#Edges & \#Time points \\
        \midrule
        PEMS04F & Flow  &\multirow{2}{*}{307} &\multirow{2}{*}{680} &\multirow{2}{*}{16992} \\
        PEMS04S & Speed  \\
        \midrule
        PEMS08F & Flow  &\multirow{2}{*}{170} &\multirow{2}{*}{548} &\multirow{2}{*}{17856}\\
        PEMS08S & Speed \\
        \bottomrule
    \end{tabular}
    \label{tab:datasets}
\end{table}

Six probabilistic baselines are considered in our experiments, including four diffusion methods: TimeGrad \cite{rasul2021autoregressive}, GCRDD \cite{ruikun2023}, DiffSTG \cite{wen2023diffstg}, and PriSTI \cite{liu2023pristi}; and two non-diffusion methods, DeepVAR \cite{salinas2020deepar} and TransNVP \cite{rasul2020multi}.  
All baselines are state-of-the-art diffusion models proposed in recent years. 
For more discussions on the baseline models, please see \textbf{Appendix C}.

\subsection{Metrics}

\noindent We use three metrics to evaluate the performance of SpecSTG, including two deterministic metrics, Root Mean Squred Error (RMSE) and Mean Absolute Error (MAE), and one probabilistic metric, Continuous Ranked Probability Score (CRPS). RMSE and MAE are adopted to measure the distance between predictions and the ground truth. We use the mean of generated samples as predictions to calculate RMSE and MAE. Given predictions $\hat{\bm{X}}_f$ at time $t_0$ (after the Fourier reconstruction) and ground truth $\bm{X}_f$ of one future window, the formulas can be written as:
\begin{align}
    \mathrm{RMSE}(\hat{\bm{X}}_f,\bm{X}_f) &= \sqrt{\frac{1}{f}\sum_{t=t_0+1}^{t_0+f} (\bm{x}_t - \hat{\bm{x}}_t)^2},\\
    \mathrm{MAE}(\hat{\bm{X}}_f,\bm{X}_f) &= \frac{1}{f}\sum_{t=t_0+1}^{t_0+f} |\bm{x}_t - \hat{\bm{x}}_t|.
\end{align}
\noindent The final results reported in our experiments are the averages from all available predictive windows in the test set.
CRPS  is a probabilistic metric that measures the compatibility of the learned probabilistic distribution at each observation ~\cite{matheson1976scoring}. Given the cumulative distribution function (CDF) $F$ of the distribution estimated at observation $x$, CRPS is defined as
\begin{equation}
    \mathrm{CRPS}(F^{-1}, x) = \int_{0}^{1} 2 \left(\kappa - \mathbb{I}_{x < F^{-1}(\kappa)}\right)\!\left(x - F^{-1}(\kappa)\right) \mathrm{d} \kappa,
\end{equation}
where $\kappa \in [0,1]$, $F^{-1}$ is the quantile function, and $\mathbb{I}_{x < F^{-1}(\kappa)}$ is an indicator function which equals to $1$ when $x < F^{-1}(\kappa)$ and $0$ otherwise. To calculate the integral, we use 100 samples generated at each time point and sensor/node to approximate the corresponding distribution and calculate CRPS following the way defined in~\cite{tashiro2021csdi}. For each future window, we may compute the normalized CRPS at time point $t = t_0+1, ..., t_0+f$ as $\frac{\sum_{n}{\mathrm{CRPS}(F_{t,n}^{-1},x_{t,n})}}{\sum_{n}|x_{t,n}|}$, 
where $n = 1, ..., N$ denotes each sensor/node, and $x_{t,n}$ is the value of sensor/node $n$ at time $t$. 
Likewise, the ``CRPS Avg.'' is computed as $\frac{\sum_{t,n}{\mathrm{CRPS}(F_{t,n}^{-1},x_{t,n})}}{\sum_{t,n}|x_{t,n}|}$.
We do not adopt $\mathrm{CRPS}_{sum}$ ~\cite{rasul2021autoregressive,tashiro2021csdi} as a metric because sensors are not regarded as features in our experiments. 
The results reported for CRPS are also averages over all available predictive windows in the test set.

\begin{table*}[t!]
\centering
\caption{The results of traffic forecasting experiments in a future window of 60 minutes. Average RMSE, MAE, CRPS, and their point values at 15/30/60 minutes are reported.  Lower values indicate better forecasting performance. The best results are marked in \textbf{bold} and the second best results are \underline{underlined}. Improvements of SpecSTG on existing methods are shown in percentage.}
\setlength{\tabcolsep}{12pt}
\scalebox{0.75}{
\begin{tabular}{@{}lllllllllllll@{}}
\toprule
\multicolumn{1}{c|}{} & \multicolumn{4}{c|}{RMSE} & \multicolumn{4}{c|}{MAE} & \multicolumn{4}{c}{CRPS} \\ \cmidrule(l){2-13} 
\multicolumn{1}{c|}{\multirow{-2}{*}{Models}} & \multicolumn{1}{c}{Avg.} & \multicolumn{1}{c}{15min} & \multicolumn{1}{c}{30min} & \multicolumn{1}{c|}{60min} & \multicolumn{1}{c}{Avg.} & \multicolumn{1}{c}{15min} & \multicolumn{1}{c}{30min} & \multicolumn{1}{c|}{60min} & \multicolumn{1}{c}{Avg.} & \multicolumn{1}{c}{15min} & \multicolumn{1}{c}{30min} & \multicolumn{1}{c}{60min} \\ \midrule
\multicolumn{13}{c}{\cellcolor[HTML]{EFEFEF}PEMS04F} \\ \midrule
DeepVAR & 50.59 & 43.90 & 48.76 & \multicolumn{1}{l|}{60.46} & 37.74 & 32.97 & 36.72 & \multicolumn{1}{l|}{45.87} & 0.2094 & 0.1997 & 0.2108 & 0.2209 \\
TransNVP                                     & 82.74                    & 68.26                     & 81.70                      & \multicolumn{1}{l|}{99.81}          & 61.85                    & 53.06                     & 62.25                     & \multicolumn{1}{l|}{73.49}          & 0.2359                   & 0.2008                    & 0.2377                    & 0.2819                    \\
TimeGrad & 35.58 & 33.22 & 35.24 & \multicolumn{1}{l|}{38.95} & \underline{21.70} & 20.26 & \underline{21.56} & \multicolumn{1}{l|}{\underline{24.04}} & 0.0801 & 0.0747 & 0.0795 & 0.0887 \\
GCRDD & 36.28 & \underline{31.94} & 45.31 & \multicolumn{1}{l|}{41.99} & 22.16 & \underline{19.48} & 21.87 & \multicolumn{1}{l|}{26.18} & 0.0779 & \underline{0.0689} & 0.0768 & 0.0982 \\
DiffSTG & 37.62 & 34.99 & 36.68 & \multicolumn{1}{l|}{43.04} & 24.90 & 22.53 & 24.65 & \multicolumn{1}{l|}{29.24} & 0.0904 & 0.0815 & 0.0894 & 0.1077 \\
PriSTI & \underline{33.74} & 33.56 & \underline{33.71} & \multicolumn{1}{l|}{\underline{37.31}} & 22.46 & 21.65 & 22.32 & \multicolumn{1}{l|}{25.19} & \underline{0.0772} & 0.0751 & \underline{0.0764} & \underline{0.0870} \\ \midrule
SpecSTG & \textbf{33.15} & \textbf{30.07} & \textbf{32.81} & \multicolumn{1}{l|}{\textbf{37.29}} & \textbf{21.53} & \textbf{19.29} & \textbf{21.39} & \multicolumn{1}{l|}{\textbf{23.29}} & \textbf{0.0766} & \textbf{0.0683} & \textbf{0.0761} & \textbf{0.0866} \\ 
Improve. & 1.75\% & 5.86\% & 2.67\% & \multicolumn{1}{l|}{0.54\%} &0.78\% &0.98\% &0.79\% &\multicolumn{1}{l|}{3.12\%} & 0.78\% &0.87\% &0.39\% & 0.46\%\\
\midrule
\multicolumn{13}{c}{\cellcolor[HTML]{EFEFEF}PEMS08F} \\ \midrule
DeepVAR & 41.43 & 35.83 & 39.88 & \multicolumn{1}{l|}{49.41} & 27.86 & 23.19 & 27.03 & \multicolumn{1}{l|}{34.89} & 0.1291 & 0.1219 & 0.1269 & 0.1412 \\
TransNVP  & 67.69 & 60.48& 68.48 & \multicolumn{1}{l|}{76.16} & 51.37 & 49.08 & 52.31                     & \multicolumn{1}{l|}{58.38}          & 0.1802                   & 0.1601                    & 0.1822                    & 0.2073                    \\
TimeGrad & 33.09 & 30.17 & 32.53 & \multicolumn{1}{l|}{37.51} & 20.47 & 18.24 & 20.06 & \multicolumn{1}{l|}{24.24} & 0.0705 & 0.0618 & 0.0695 & 0.0843 \\
GCRDD & 28.83 & \underline{23.91} & 28.10 & \multicolumn{1}{l|}{35.68} & 18.72 & \underline{15.52} & 18.35 & \multicolumn{1}{l|}{23.99} & 0.0626 & \underline{0.0517} & 0.0617 & 0.0833 \\
DiffSTG & 28.26 & 25.04 & 27.54 & \multicolumn{1}{l|}{34.32} & 18.99 & 16.66 & 18.63 & \multicolumn{1}{l|}{23.68} & 0.0692 & 0.0609 & 0.0679 & 0.0872 \\
PriSTI & \underline{26.35} & 24.58 & \underline{26.93} & \multicolumn{1}{l|}{\underline{29.91}} & \underline{17.30} & 15.98 & \underline{17.32} & \multicolumn{1}{l|}{\underline{20.67}} & \underline{0.0576} & 0.0539 & \underline{0.0581} & \underline{0.0688} \\ 
\midrule
SpecSTG & \textbf{25.59} & \textbf{22.23} & \textbf{24.77} & \multicolumn{1}{l|}{\textbf{29.90}} & \textbf{17.06} & \textbf{14.93} & \textbf{16.70} & \multicolumn{1}{l|}{\textbf{20.25}} & \textbf{0.0572} & \textbf{0.0500} & \textbf{0.0558} & \textbf{0.0680} \\ 
Improve. & 2.88\% & 7.03\% & 8.02\% & \multicolumn{1}{l|}{0.03\%} &1.39\% &3.80\% &3.58\% &\multicolumn{1}{l|}{2.03\%} & 0.69\% &3.29\% &3.96\% & 1.16\% \\
\midrule
\multicolumn{13}{c}{\cellcolor[HTML]{EFEFEF}PEMS04S} \\ \midrule
DeepVAR & 6.23 & 5.72 & 6.11 & \multicolumn{1}{l|}{6.93} & 2.76 & 2.52 & 2.72 & \multicolumn{1}{l|}{3.13} & 0.0490 & 0.0450 & 0.0488 & 0.0549 \\
TransNVP                                     & 6.25                     & 5.73                      & 6.27                      & \multicolumn{1}{l|}{6.98}           & 3.36                     & 3.12                      & 3.39                      & \multicolumn{1}{l|}{3.74}           & 0.0408                   & 0.0382                    & 0.0412                    & 0.0446                    \\
TimeGrad & 5.92 & 5.62 & 5.91 & \multicolumn{1}{l|}{6.35} & 2.38 & 2.19 & 2.37 & \multicolumn{1}{l|}{2.66} & 0.0307 & 0.0282 & 0.0308 & 0.0345 \\
GCRDD & \underline{4.33} & \underline{3.10} & \underline{4.30} & \multicolumn{1}{l|}{5.63} & \underline{1.94} & \underline{1.51} & \textbf{1.97} & \multicolumn{1}{l|}{\underline{2.58}} & \textbf{0.0245} & \textbf{0.0189} & \textbf{0.0248} & \underline{0.0329} \\
DiffSTG & 4.46 & 3.24 & 4.46 & \multicolumn{1}{l|}{5.72} & 2.15 & 1.66 & 2.20 & \multicolumn{1}{l|}{2.83} & 0.0264 & 0.0206 & 0.0267 & 0.0340 \\
PriSTI & 4.42 & 3.31 & 4.67 & \multicolumn{1}{l|}{\underline{5.60}} & 1.96 & 1.54 & \underline{1.99} & \multicolumn{1}{l|}{2.62} & \underline{0.0252} & 0.0198 & 0.0258 & \underline{0.0329} \\
\midrule
SpecSTG & \textbf{4.06} & \textbf{3.01} & \textbf{4.09} & \multicolumn{1}{l|}{\textbf{5.15}} & \textbf{1.93} & \textbf{1.50} & \textbf{1.97} & \multicolumn{1}{l|}{\textbf{2.51}} & 
\textbf{0.0245} & \underline{0.0192} & \underline{0.0253} & \textbf{0.0319}\\ 
Improve. & 6.24\% & 2.90\% & 4.88\% & \multicolumn{1}{l|}{8.04\%} &0.52\% &0.66\% &0.00\% &\multicolumn{1}{l|}{2.71\%} & 0.00\% &- &- & 3.04\% \\
\midrule
\multicolumn{13}{c}{\cellcolor[HTML]{EFEFEF}PEMS08S} \\ \midrule
DeepVAR & 5.73 & 5.55 & 5.70 & \multicolumn{1}{l|}{6.05} & 2.56 & 2.42 & 2.57 & \multicolumn{1}{l|}{2.79} & 0.0544 & 0.0534 & 0.0543 & 0.0558 \\
TransNVP                                     & 5.41                     & 5.12                      & 5.54                      & \multicolumn{1}{l|}{5.74}           & 2.76                     & 2.64                      & 2.83                      & \multicolumn{1}{l|}{2.91}           & 0.0349                   & 0.0334                    & 0.0358                    & 0.0368                    \\
TimeGrad & 4.98 & 4.93 & 4.97 & \multicolumn{1}{l|}{5.03} & 1.98 & 1.95 & 1.97 & \multicolumn{1}{l|}{\underline{2.12}} & 0.0267 & 0.0262 & 0.0268 & \underline{0.0272} \\
GCRDD & \underline{3.75} & \underline{2.74} & \underline{3.77} & \multicolumn{1}{l|}{4.89} & 1.72 & 1.35 & \underline{1.75} & \multicolumn{1}{l|}{2.32} & \underline{0.0223} & 0.0171 & \underline{0.0226} & 0.0301 \\
DiffSTG & 3.97 & 3.20 & 4.07 & \multicolumn{1}{l|}{\underline{4.82}} & 2.36 & 1.91 & 2.43 & \multicolumn{1}{l|}{3.04} & 0.0325 & 0.0259 & 0.0335 & 0.0423 \\
PriSTI & 4.22 & 3.02 & 4.39 & \multicolumn{1}{l|}{5.20} & \underline{1.70} & \underline{1.29} & 1.80 & \multicolumn{1}{l|}{2.15} & \textbf{0.0217} & \textbf{0.0162} & 0.0230 & \underline{0.0272} \\
\midrule
SpecSTG & \textbf{3.45} & \textbf{2.58} & \textbf{3.46} & \multicolumn{1}{l|}{\textbf{4.36}} & \textbf{1.63} & \textbf{1.27} & \textbf{1.67} & \multicolumn{1}{l|}{\textbf{2.02}} & \textbf{0.0217} & \underline{0.0170} & \textbf{0.0219} & \textbf{0.0268} \\ 
Improve. & 8.00\% & 5.84\% & 8.22\% & \multicolumn{1}{l|}{9.54\%} &4.12\% &1.55\% &4.57\% &\multicolumn{1}{l|}{4.72\%} & 0.00\% &- &0.10\% & 1.47\% \\
\bottomrule
\end{tabular}}
\label{tab:expresults}
\end{table*}

\subsection{Implementation Details}
We implement SpecSTG on a single NVIDIA 4090 GPU with 64GB of memory. The model is trained with the Adam optimizer with a learning rate schedule from $5e-4$ to $1e-2$. 
The maximum number of epochs is 300 for flow data and 50 for speed data with batch size 64. Validation loss is used for model selection. 
The hyperparameters specific to diffusion models are set as follows. We use the quadratic scheme for noise level $\beta_k$ starting from $\beta_1 = 1e-4$ and tune $\beta_K$ in $[0.1, 0.2, 0.3,0.4]$. The number of diffusion steps $K$ is selected from $[50,100,200]$. The maximum polynomial order in $\mathrm{SpecConv}$ is set as $2$. The hidden size $D_h$ is tuned in $[64, 96]$. In $\text{SG-Wave}$, the number of residual blocks $M = 8$ and the residual channel $D_r = 8$. Finally, the number of samples $S$ is set as 100 for all models. For all experiments, we split datasets with 60\%/20\%/20\%  train/validation/test proportions and apply Z-score normalization before the Fourier transform.
The graph structure is constructed depending on the distance between sensors following~\cite{guo2019attention}. More implementation details are presented in \textbf{Appendix D}.

\begin{figure*}[t]
\begin{minipage}[b]{1.0\linewidth}
  \centering
  \centerline{\includegraphics[height = 8.5cm, width=17.7cm]{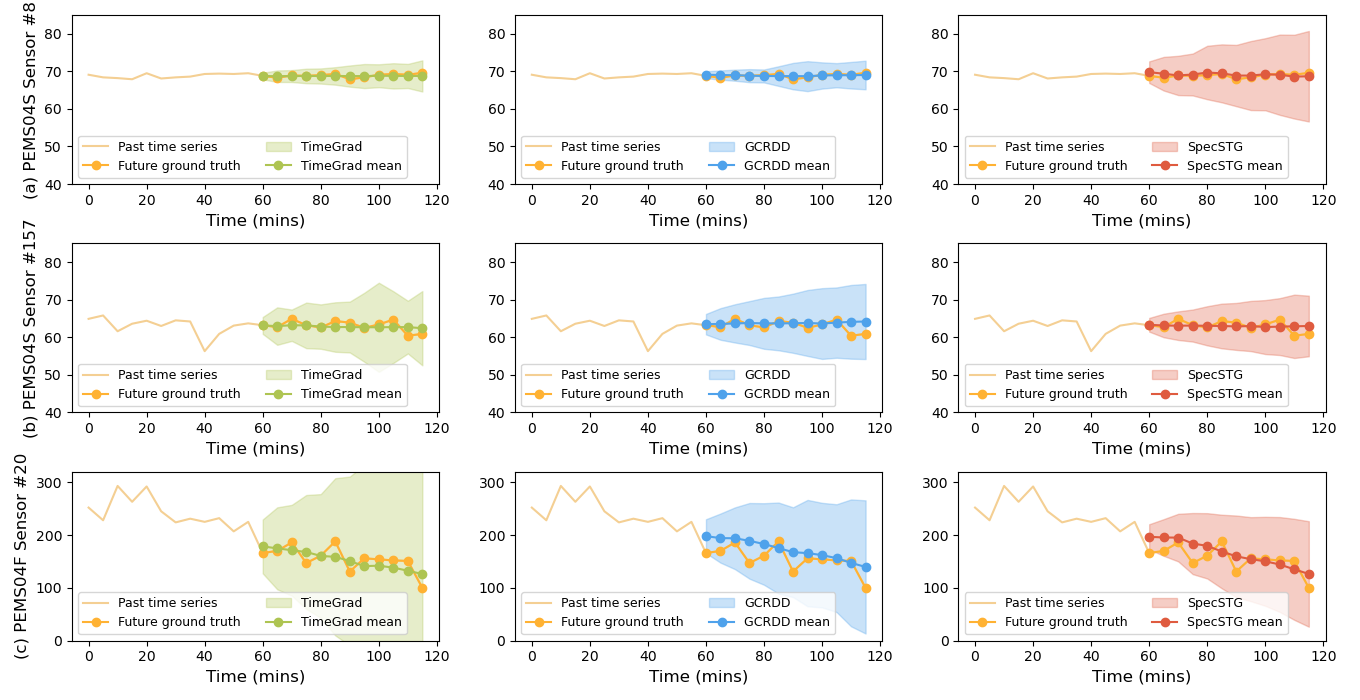}}
\end{minipage}
\caption{Forecasting visualizations of TimeGrad (green), GCRDD (blue), and SpecSTG (red). (a) and (b) are results on speed data (PEMS04S), while (c) presents results on flow data (PEMS04F).}
\label{fig:forecast}
\end{figure*}

\subsection{Experiment Results} \label{sec:exp_analysis}
\noindent \textbf{Table \ref{tab:expresults}} presents the results of our traffic forecasting experiments, where the prediction task involves forecasting future time series for a 60-minute horizon based on observations from the past 60 minutes. The table includes numerical values for the average RMSE, MAE, and CRPS over the entire forecast window. Additionally, it provides corresponding point evaluations assessed at 15, 30, and 60 minutes such that we can evaluate the short and long-term forecasting performance of the models. 

\noindent\textbf{Analysis of deterministic results} SpecSTG consistently achieves top-tier deterministic results across various tasks. 
In comparison to the second-best model, SpecSTG exhibits an 8.00\% improvement in average RMSE for PEMS08S and a 6.24\% improvement for PEMS04S. Similarly, the average MAE sees enhancements of 4.12\% for PEMS08S and 1.39\% for PEMS08F. In addition, our method shows proficiency in both short and long-term deterministic forecasting. Particularly, the RMSE improvement of SpecSTG achieves 7.03\% on PEMS08F for short-term 15-minute forecasting and 9.54\% on PEMS08S for long-term 60-minute forecasting. We also observe improvements in MAE for most results in the table, especially for traffic flow tasks. The superior deterministic performance stems from SpecSTG's unique probabilistic learning process in the spectral domain, leveraging rich global spatial information, which is an acknowledged crucial component in deterministic traffic forecasting \cite{yu2017spatio,fang2019gstnet}. 

\noindent\textbf{Analysis of probabilistic results} Regarding the probabilistic metric CRPS, SpecSTG demonstrates an advantage in traffic flow forecasting but not in vehicle speed forecasting. The average CRPS is improved by 0.78\% and 0.69\% with SpecSTG on PEMS04F and PEMS08F, respectively. However, little improvement is observed on PEMS04S and PEMS08S. Since speed and flow datasets share the same graph structure (e.g., PEMS04S and PEMS04F), we suggest that the inferior probabilistic performance is related to the time series data. A possible explanation is that flow data is often associated with higher variations, thus the systematic variations measured by the Fourier representation are more informative than the speed data. For instance, the standard deviations of data in PEMS08S and PEMS08F are 6.65 and 146.22, respectively. We will further investigate this observation with forecasting visualizations in Subsection \ref{sec:visual}.


\noindent\textbf{Comments on baseline models} Non-diffusion models such as DeepVAR and TransNVP clearly show inferior performance compared to diffusion models. PriSTI excels in predicting traffic flow, leveraging the attention mechanism to process temporal dynamics. On the other hand, TimeGrad and GCRDD exhibit comparably better performance in forecasting vehicle speeds, utilizing a recurrent structure that emphasizes consecutive temporal connectivity. We suppose this distinction is caused by the attention mechanism's effectiveness in capturing global temporal variations and the recurrent structure's aptness for learning highly correlated speed patterns. Our approach, with a recurrent encoder, combines the strengths of both methodologies. It preserves the ability to learn continuous time patterns while enhancing the identification of variations through the incorporation of spectral variation measurements.

\section{\textsc{DISCUSSIONS}}

In this section, we will visualize SpecSTG's forecasting outcomes compare with two baseline diffusion models. In addition, we will provide analyses on time efficiency and sensitivity to hyperparameters to further show the advantage of SpecSTG. More supplementary discussions can be found in \textbf{Appendices E}, \textbf{F}, and \textbf{G}.

\subsection{Forecasting Visualizations}\label{sec:visual}

\noindent Recall that in Subsection \ref{sec:exp_analysis}, SpecSTG performs better in probabilistic forecasting on traffic flow data than speed data. Here we further explore this observation by visualizing the forecasting outcomes of TimeGrad, GCRDD, and SpecSTG on PEMS04S and PMES04F (\textbf{Figure \ref{fig:forecast}}). The figure displays the mean and $95\%$ confidence interval (adjusted by time) of estimated future distributions. Our primary objective is to assess whether the forecasting intervals produced by various methods are compatible with the actual future time series. An appropriate interval should capture the future variations while remaining sufficiently narrow to provide meaningful insights.

In traffic speed forecasting, SpecSTG's mean estimation is closer to future time series, but the intervals generated by TimeGrad and GCRDD sometimes better fit the variations in future values. Upon closer examination of the data patterns, we observe that this impact is particularly pronounced in windows with very small variations (\textbf{Figure \ref{fig:forecast} (a)}). In contrast, distributions estimated by SpecSTG at sensors with larger variations (\textbf{Figure \ref{fig:forecast} (b)}) exhibit a better ability to capture future uncertainty compared to TimeGrad and GCRDD. This matches our previous hypothesis in Subsection \ref{sec:exp_analysis} that \textbf{STG forecasting with larger variations benefits more from the spectral diffusion process}. Because more systematic fluctuations exist in such data, and thus more information can be captured by the Fourier representation, and eventually learned by the diffusion process.

Analogously, in traffic flow forecasting on PEMS04F, a dataset characterized by high systematic variation, SpecSTG demonstrates promising performance by generating both more accurate deterministic predictions and more compatible distributions (\textbf{Figure \ref{fig:forecast} (c)}). This observation reflects the experiment results that SpecSTG achieves outstanding performance with PEMS04F in terms of all metrics.

\subsection{Time Efficiency}
\begin{figure}[t!]
\begin{minipage}[b]{1.0\linewidth}
  \centering
  \centerline{\includegraphics[height = 6.5cm, width=8.5cm]{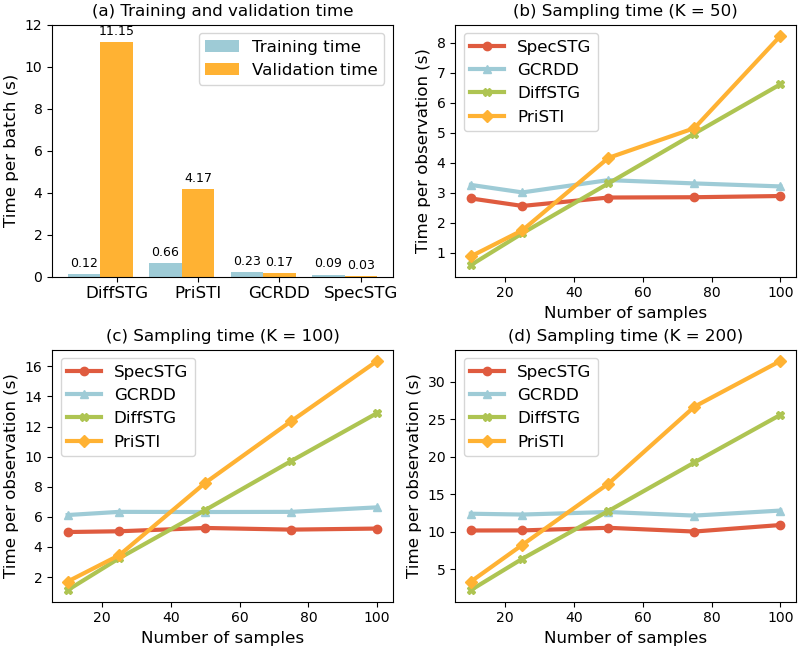}}
\end{minipage}
\caption{Time efficiency of training, validation, and sampling.}
\label{fig:time}
\end{figure}

In \textbf{Figure \ref{fig:time}}, we compare the training, validation, and sampling time of SpecSTG with three diffusion models for STG forecasting, including GCRDD, DiffSTG, and PriSTI. The training and validation of SpecSTG are clearly faster than other diffusion baselines. Particularly, SpecSTG's training plus validation time is 3.33$\times$ of GCRDD, the most efficient method among other existing state-of-the-arts. The validation time of DiffSTG is significantly high because it requires sampling and prediction during validation. To show sampling efficiency, we plot the sampling time per observation (i.e., a future window of 60 minutes) when diffusion steps $K = 50, 100, 200$. We set the batch size of one-shot methods, DiffSTG and PriSTI, as $8$ and $16$, and report the best results. For autoregressive methods, SpecSTG shows a notable time advantage over GCRDD in sampling. Besides, their sampling time does not vary much with the number of samples $S$. By contrast, the time cost of one-shot methods increases rapidly with the increase of $S$. Although DiffSTG and PriSTI are more efficient when $S$ is small, a small number of samples often cannot present a clear picture of future data distribution. 

\begin{figure}[t!]
\begin{minipage}[b]{1.0\linewidth}
  \centering
  \centerline{\includegraphics[height = 2.5cm, width=9cm]{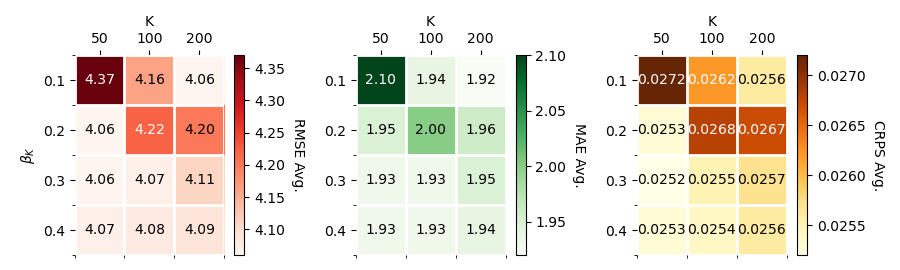}}
\end{minipage}
\caption{Sensitivity analysis of SpecSTG on key hyperparameters: diffusion steps $K$ and the end of beta schedule $\beta_K$.}
\label{fig:hyper}
\end{figure}

\subsection{Sensitivity Analysis on Hyperparameters}\label{sec:sens_analysis}
\noindent In this sensitivity analysis, we focus on the combination of two very important diffusion hyperparameters: the number of diffusion steps, $K$, and the end of noise schedule, $\beta_K$. Theoretically, the choices of $K$ and $\beta_K$ are relevant to each other. Since the noise level gradually increases from $\beta_1 = 1e-4$ to $\beta_K$ in $K$ diffusion steps, these two hyperparameters control the changing speed and level of noises in the diffusion process, which is essential for the white noise assumption of diffusion models. 
The same as the implementation of SpecSTG in our experiments, we set the search spaces of $\beta_K$ and $K$ as $[0.1, 0.2, 0.3, 0.4]$ and $[50, 100, 200]$, respectively. In \textbf{Figure \ref{fig:hyper}}, we use heatmaps to show the change in model performance in terms of RMSE, MAE, and CRPS on PEMS04S with different hyperparameter combinations. We observe that SpecSTG typically performs better with a larger $\beta_K$ (for instance $0.3$ or $0.4$). The best results appear when $\beta_K = 0.3$ and $K = 50$. It is also worth noting that the forecasting performance of SpecSTG does not vary dramatically with different hyperparameter combinations. This means our method is not very sensitive to hyperparameter selection, alleviating the burden of hyperparameter tuning.

\section{\textsc{RELATED WORKS}}
\textbf{Generative Diffusion Models} The initial idea of diffusion models was introduced by \cite{sohl2015deep}. Then, some improvements proposed by \cite{ho2020denoising} endowed them with remarkable practical value, contributing to their conspicuous popularity nowadays. 
In recent years, diffusion models have demonstrated their power over many existing generative techniques in various real-world applications such as image synthesis \cite{austin2021structured,dhariwal2021diffusion,ho2022cascaded}, video generation \cite{harvey2022flexible,ho2022video,yang2022video}, natural language processing \cite{li2022diffusion,savinov2022step,yu2022latent}, and time series prediction \cite{rasul2021autoregressive,li2022generative,lopezalcaraz2023diffusionbased}. 

\noindent\textbf{Diffusion Models for Time Series and STGs} Pioneering diffusion models for time series such as TimeGrad \cite{rasul2021autoregressive} and TimeDiff \cite{shen2023non} were originally tailored for multivariate time series forecasting, utilizing sensors as variables but only exploring general dependencies without incorporating graph structural information. 
Recently, several diffusion models were proposed specifically for STG forecasting. DiffSTG \cite{wen2023diffstg} incorporates graph structure into the backward kernel with a graph-modified Unet \cite{ronneberger2015u} architecture. GCRDD \cite{ruikun2023}, designed in reminiscent of TimeGrad, adopts a graph-enhanced recurrent encoder to produce hidden states from past time series as conditions. Additionally, USTD \cite{hu2023towards} introduces a pre-trained encoder that better captures deterministic patterns via an unsupervised reconstruction task. DVGNN \cite{liang2023dynamic} is a deterministic model but with a diffusion module to generate dynamic adjacency matrices in its pre-training process. Furthermore, PriSTI \cite{liu2023pristi} was initially developed from CSDI \cite{tashiro2021csdi} for STG imputation, 
but with potential for forecasting tasks by masking future data as missing values. We highlight that SpecSTG's novelty lies in its unique spectral diffusion framework that generates graph Fourier representation of future time series, which leverages systematic fluctuations in time series data guided by graph structure to boost forecasting accuracy. 

\noindent\textbf{Spectral diffusion on graphs and time series}
The idea of spectral diffusion has been applied in generating graph structure and classic time series. GSDM \cite{luo2023fast} explores the generation of spectral graph structure, i.e., the eigenvalues of graph adjacency matrices, to enhance graph generation quality. Besides, research has shown that generating classic time series in the Fourier domain facilitates diffusion models to better capture the training distribution \cite{crabbe2024time}. Our method, SpecSTG, is the first endeavour to investigate spectral diffusion in generating graph signals and spatio-temporal data.

\section{\textsc{CONCLUDING REMARK}}
In this paper, we proposed SpecSTG, a spectral diffusion approach for fast probabilistic spatio-temporal traffic forecasting. Our method transforms the entire diffusion learning process to the spectral domain by generating the Fourier representation of future time series instead of the original data. Although we have introduced the autoregressive architecture of SpecSTG, the idea of spectral diffusion can be straightforwardly applied to one-shot methods as well by altering the generative target and graph convolution. Hence, SpecSTG can be regarded as an effective framework for STG forecasting. Experiment results confirm the superior performance of SpecSTG, demonstrating more efficient training and sampling compared to state-of-the-art diffusion methods. Nevertheless, we highlight that SpecSTG may fall short of predicting compatible future distributions when the data have low variations, diminishing the efficacy of spectral measurements of systematic fluctuations.

\bibliographystyle{ACM-Reference-Format}
\bibliography{reference.bib}

\clearpage
\section*{APPENDICES}

\section*{Appendix A. Denoising Diffusion Probabilistic Model}\label{apx:ddpm}
Denoising diffusion probabilistic model (DDPM) is one of the classic formulations of diffusion models ~\cite{sohl2015deep,ho2020denoising,yang2023diffusion}. With its high flexibility in modelling target distributions and outstanding capability of capturing complex generative patterns, DDPM is widely used in various time series tasks ~\cite{rasul2021autoregressive,tashiro2021csdi,coletta2023constrained,lin2023diffusion}.
Unlike traditional probabilistic models that learn the target distribution $q(x)$ explicitly to generate samples, DDPM learns how to generate samples directly via a pair of forward-backward Markov chains without accessing the actual target distribution function. Assuming that $\bm{x}^0$ is the original data, for $k = 0, 1, ..., K$, the forward chain injects Gaussian noises to the generative target with a probabilistic transition kernel 
\begin{equation*}
    q(\bm{x}^k|\bm{x}^{k-1}) = \mathcal{N}\left( \bm{x}^k; \sqrt{1 -\beta_k} \bm{x}^{k-1}, \beta_k \bm{I} \right),
\end{equation*}
where $\beta_k \in (0,1)$ is a hyperparameter controlling the noise level at each forward step. The design of forward chain enables us to derive the disturbed data at a particular step $k$ directly from $\bm{x}^0$ as
\begin{equation*}\label{eq_forwardproperty}
    q(\bm{x}^k|\bm{x}^{0}) = \mathcal{N}\left( \bm{x}^k; \sqrt{\Tilde{\alpha}_k} \bm{x}^{0}, (1-\Tilde{\alpha}_k) \bm{I} \right),
\end{equation*}
where $\Tilde{\alpha}_k \coloneqq \prod_{i=1}^k {(1-\beta_i)}$. This means $\bm{x}^k = \sqrt{\Tilde{\alpha}_k} \bm{x}^0 + \sqrt{1-\Tilde{\alpha}_k} \bm{\epsilon}$ with $\bm{\epsilon} \sim \mathcal{N}(\bm{0}, \bm{I})$. To ensure that the forward chain transits eventually to white noises, one shall have $\Tilde{\alpha}_K \approx 0$ such that $q(\bm{x}^K) \coloneqq \int q(\bm{x}^K|\bm{x}^0) q(\bm{x}^0) \mathrm{d}\bm{x}^0 \approx \mathcal{N}(\bm{x}^K; 0, \bm{I})$. 
Next, the backward chain recovers white noises to original data through a Gaussian transition kernel $p_{\bm{\theta}}(\bm{x}^{k-1}|\bm{x}^{k}) = \mathcal{N}\left( \bm{x}^{k-1};\bm{\mu}_{\bm{\theta}}(\bm{x}^k,k),
    \sigma_k \mathbf{I}\right)$,
where $\bm{\mu}_{\bm{\theta}}(\bm{x}^k,k)$ is usually parameterized with a neural network with learnable parameters $\bm{\theta}$, and $\sigma_k$ is a variance hyperparameter.
These parameters are optimized by minimizing the negative evidence lower-bound (ELBO):
\begin{equation*}
    \mathcal{L}_{E}(\bm{\theta}) = \E_{q\left(\bm{x}^{0:K}\right)} \left[ -\log{p(\bm{x}^K)} - \sum_{k=1}^K \log{\frac{p_{\bm{\theta}}(\bm{x}^{k-1}|\bm{x}^k)}{q(\bm{x}^k|\bm{x}^{k-1})}} \right].
\end{equation*}
The learning process is simplified with DDPM by \cite{ho2020denoising}. Instead of learning $\bm{\mu}_{\bm{\theta}}(\bm{x}^k,k)$, a denoising network $\bm{\epsilon}_{\bm{\theta}} \left(\bm{x}^k, k \right)$ is learned with the following objective function:
\begin{equation*}
    \mathcal{L}_{DDPM}(\bm{\theta}) = \E_{k, \bm{x}^0, \bm{\epsilon}}  \left\|\bm{\epsilon} - \bm{\epsilon}_{\bm{\theta}} \left(\bm{x}^k, k \right)\right\|^2.
\end{equation*}
This denoising network $\bm{\epsilon}_{\bm{\theta}} (\bm{x}^k, k)$ takes noised data $\bm{x}^k$ and step index $k$ as inputs to predict the noise injected at step $k$ in the forward chain. Finally, DDPM generates samples by eliminating the noises in random white noise $\bm{x}^K \sim \mathcal{N}(\bm{x}^K; 0, \bm{I})$. For backward step $k = K, K-1, ..., 1$, DDPM updates the sample as
\begin{equation*}
    \bm{x}^{k-1} \leftarrow \frac{1}{\sqrt{1-\beta_{k}}} 
    \left(
    \bm{x}^{k} - \frac{\beta_{k}}{\sqrt{1 - \Tilde{\alpha}_{k}}}
    \bm{\epsilon}_{\bm{\theta}} (\bm{x}^{k}, k)
    \right) + \sigma_{k} \bm{e}
\end{equation*}
where $\sigma_k$ is a hyperparameter of variance in the backward transition kernel with $\bm{e} = \bm{0}$ for $k = 1$, and $\bm{e} \sim \mathcal{N}(\bm{0},\bm{I})$ otherwise. Eventually, $\bm{x}^0$ will be a sample from the same distribution of the generative target.

\section*{Appendix B. Generalized Fourier Transform}\label{apx:gen_fourier_transform}
The Fourier transform can be naturally generalized to multivariate traffic STGs. Assume that we have a traffic STG with graph signals $\bm{X}_{\mathcal{G}} = \{\bm{x}_1, \bm{x}_2, ..., \bm{x}_t, ...| \bm{x}_t \in \mathbb{R}^{N \times D_x}\}$, where $D_x$ is the number of variables. The graph is a univariate STG when $D_x = 1$, and a multivariate STG when $D_x \geq 2$.
For a sampled past-future window $\bm{X} = \{\bm{x}_{t_0-c+1}, ..., \bm{x}_{t_0+f}\} \in \mathbb{R}^{N \times D_x \times (c+f)}$, we first split it according to its variables and rewrite it as $\bm{X} = \{\bm{x}^\prime_1, ...,\bm{x}^\prime_{d},...,\bm{x}^\prime
_{D_x}| \bm{x}^\prime_d \in \mathbb{R}^{N \times (c + f)}\}$. 
Then, with the Fourier operator $\bm{U}$, we conduct the transform on each variable matrix $\bm{x}^\prime_d$ as
\begin{equation*}
    \Tilde{\bm{x}}^\prime_d = \bm{U}^\intercal \bm{x}^\prime_d.
\end{equation*}
Hence, the Fourier representation of the sampled window $\bm{X}$ is $\Tilde{\bm{X}} = \{\Tilde{\bm{x}}^\prime_1, ..., \Tilde{\bm{x}}^\prime_d, ..., \Tilde{\bm{x}}^\prime_{D_x}\} \in \mathbb{R}^{N \times D_x \times (c+f)}$. To convert the Fourier representation of a single variable matrix back to the original domain, we may apply Fourier reconstruction as
\begin{equation*}
    \bm{x}^\prime_{d} = \bm{U} \Tilde{\bm{x}}^\prime_{d},
\end{equation*}
so $\bm{X} = \{\bm{U} \Tilde{\bm{x}}^\prime_{1}, ..., \bm{U} \Tilde{\bm{x}}^\prime_{d}, ..., \bm{U} \Tilde{\bm{x}}^\prime_{D_x}\} \in \mathbb{R}^{N \times D_x \times (c+f)}$.

\section*{Appendix C. Baselines}\label{apx:baselines}
\noindent In the main experiment, we compare SpecSTG with four state-of-the-art diffusion baselines:

\begin{itemize}
    \item \textbf{TimeGrad} \cite{rasul2021autoregressive}: an autoregressive diffusion model using long short-term memory (LSTM) or gated recurrent units (GRU) to encode temporal dynamics and a time-modified WavaNet \cite{oord2016wavenet} as $\bm{\epsilon}_{\bm{\theta}}$.
    

    \item \textbf{GCRDD} \cite{ruikun2023}: an autoregressive diffusion model for spatio-temporal forecasting, which adopts a graph-modified GRU to encode past time series and spatial connectivity as conditions. Developed from TimeGrad, GCRDD also employs a WaveNet architecture for its $\bm{\epsilon}_{\bm{\theta}}$ but with graph convolution to process spatial information.
    
    \item \textbf{DiffSTG} \cite{wen2023diffstg}: an one-shot diffusion model for spatio-temporal forecasting with graph-modified UNet \cite{ronneberger2015u} as $\bm{\epsilon}_{\bm{\theta}}$.

    \item \textbf{PriSTI} \cite{liu2023pristi}: a one-shot diffusion model for spatio-temporal imputation. It is equipped with a feature extraction mechanism to construct conditions. An attention-based  \cite{vaswani2017attention} WaveNet is adopted as its $\bm{\epsilon}_{\bm{\theta}}$.
\end{itemize}
In addition, we also compare SpecSTG with two non-diffusion probabilistic methods, including
\begin{itemize}

    \item \textbf{DeepVAR} \cite{salinas2020deepar}: an autoregressive RNN-based model for multivariate time series forecasting, known as the multivariate variant of DeepAR.

    \item \textbf{Transformer Normalizing Flow} \cite{rasul2020multi}: an autoregressive model for multivariate time series forecasting that approximates the target distribution with a normalizing flow such as Real-NVP \cite{dinh2017density} and MAF \cite{papamakarios2017masked}. Here we choose Real-NVP, and we call this model TransNVP.

\end{itemize}
In addition, in \textbf{Appendix E}, we provide supplementary comparisons with non-diffusion probabilistic methods, including
\begin{itemize}
    \item \textbf{Historical Average (HA)}: a deterministic method that regards time series as periodic processes and predicts future time series with weighted averages from past observations.

    \item \textbf{Vector Auto-Regressive model (VAR)} \cite{zivot2006vector}: a deterministic model for multivariate time series forecasting that assumes time series are stationary and predicts with lagged observations.

    \item \textbf{FC-LSTM} \cite{sutskever2014sequence}: an LSTM model with fully connected (FC) hidden units that performs well in capturing sequential temporal dependencies. 

    \item \textbf{STGCN} \cite{yu2017spatio}: a graph convolutional network for spatio-temporal traffic forecasting, equipped with gated temporal convolution and graph convolution to process temporal and spatial information.

    \item \textbf{DCRNN} \cite{li2017diffusion}: a recurrent graph neural network that integrates graph diffusion and GRU under an encoder-decoder structure.

    \item \textbf{ASTGCN} \cite{guo2019attention}: an attention-based graph convolutional neural network that adopts both graph attention and temporal attention to learn spatio-temporal patterns.

    \item \textbf{GMAN} \cite{zheng2020gman}: a graph neural network with multiple spatio-temporal attention blocks in its encoder-decoder architecture to enhance the learning of spatio-temporal patterns.
\end{itemize}

\section*{Appendix D. Implementation Details}\label{apx:imp_details}
\textbf{Diffusion baselines} Diffusion models are implemented with similar diffusion process hyperparameters as SpecSTG. For all models that utilize WaveNet architecture in their denoising networks, we fix the number of residual blocks and residual channels both as $8$. Other model-specific implementation details follow their original papers along with the default settings in their codes.

\noindent\textbf{Non-diffusion probabilistic baselines} DeepVAR is implemented with \texttt{PyTorchTS}\footnote{Source: \url{https://github.com/zalandoresearch/pytorch-ts}}. We choose the LSTM with 2 layers as its recurrent structure. The size of hidden states is fixed at $1024$, because a smaller hidden size leads to unsatisfactory performance, and a larger hidden size will not improve the results much. TransNVP is also implemented with \texttt{PyTorchTS}. Given that details of TransNVP are not mentioned in its paper, we set its hyperparameters majorly according to the settings of Transformer MAF, which is a similar model but uses MAF to model the target distribution. We also tune the hyperparameters when needed to improve the performance of TransNVP.

\noindent\textbf{Deterministic baselines} The results of deterministic baselines are retrieved directly from the experiments in  \cite{shao2022decoupled}, which adopts similar experiment settings (e.g., 60\%/20\%/20\% data split) as our experiments.

\section*{Appendix E. Comparison with Classic Deterministic Baselines}\label{apx:deter}

\textbf{Table \ref{tab:deter}} presents a comparison between SpecSTG and deterministic methods. Notably, SpecSTG outperforms most classic deterministic approaches, including traditional statistical models like VAR and graph neural networks such as DCRNN and ASTGCN. However, GMAN consistently demonstrates superior performance, particularly in long-term forecasting. Here are two plausible explanations. Firstly, being a deterministic model, GMAN is trained with MAE loss, strengthening the generation of accurate deterministic predictions compared to SpecSTG, which optimizes an implicit probabilistic objective. Secondly, GMAN's sophisticated encoder-decoder architecture and multiple spatio-temporal attention blocks effectively integrate spatial and temporal information, while SpecSTG relies on a single graph-modified GRU encoder for processing spatio-temporal patterns. This observation could guide future work to enhance our model.

\begin{table}[t!]
\centering
\caption{Comparison between SpecSTG and classic deterministic models on PEMS04F and PEMS08F. Partial results are retrieved from~\protect\cite{shao2022decoupled}. The best results are marked in \textbf{bold}, and the second best results are {\underline {underlined}}.}
\setlength{\tabcolsep}{12pt}
\renewcommand{\arraystretch}{1}
\scalebox{0.68}{
\begin{tabular}{@{}lrrrrrr@{}}
\toprule
\multicolumn{1}{c|}{}                        & \multicolumn{3}{c|}{RMSE}                                                          & \multicolumn{3}{c}{MAE}                                                           \\ \cmidrule(l){2-7} 
\multicolumn{1}{c|}{\multirow{-2}{*}{Model}} & \multicolumn{1}{c}{15min} & \multicolumn{1}{c}{30min} & \multicolumn{1}{c|}{60min} & \multicolumn{1}{c}{15min} & \multicolumn{1}{c}{30min} & \multicolumn{1}{c}{60min} \\ \midrule
\multicolumn{7}{c}{\cellcolor[HTML]{EFEFEF}PEMS04F}                                                                                                                                                                   \\ \midrule
HA                                           & 42.69                     & 49.37                     & \multicolumn{1}{r|} {67.43}                      & 28.92                     & 33.73                     & 46.97                     \\
VAR                                          & 34.30                     & 36.58                     & \multicolumn{1}{r|} {40.28}                      & 21.94                     & 23.72                     & 26.76                     \\
FC-LSTM                                      & 33.37                     & 39.1                      & \multicolumn{1}{r|} {50.73}                      & 21.42                     & 25.83                     & 36.41                     \\
STGCN                                        & 30.76                     & 34.43                     & \multicolumn{1}{r|} {41.11}                      & 19.35                     & 21.85                     & 26.97                     \\
DCRNN                                        & 31.94                     & 36.15                     & \multicolumn{1}{r|} {44.81}                      & 20.34                     & 23.21                     & 29.24                     \\
ASTGCN                                       & 31.43                     & 34.34                     & \multicolumn{1}{r|} {40.02}                      & 20.15                     & 22.09                     & 26.03                     \\
GMAN                                         & \textbf{29.32}            & \textbf{30.77}            & \multicolumn{1}{r|} {\textbf{30.21}}             & \textbf{18.28}            & \textbf{18.75}            & \textbf{19.95}            \\
\midrule
SpecSTG                                      & {\underline {30.17}}               & {\underline {32.81}}               & \multicolumn{1}{r|} {{\underline {37.29}}}                & {\underline {19.29}}               & {\underline {21.39}}               & {\underline {23.29}}               \\ \midrule
\multicolumn{7}{c}{\cellcolor[HTML]{EFEFEF}PEMS08F}                                                                                                                                                                   \\ \midrule
HA                                           & 34.96                     & 40.89                     & \multicolumn{1}{r|} {56.74}                      & 23.52                     & 27.67                     & 39.28                     \\
VAR                                          & 29.73                     & 30.30                     & \multicolumn{1}{r|} {38.97}                      & 19.52                     & 22.25                     & 26.17                     \\
FC-LSTM                                      & 26.27                     & 34.53                     & \multicolumn{1}{r|} {47.03}                      & 17.38                     & 21.22                     & 30.69                     \\
STGCN                                        & 25.03                     & 27.27                     & \multicolumn{1}{r|} {34.21}                      & 15.30                     & 17.69                     & 25.46                     \\
DCRNN                                        & 25.48                     & 27.63                     & \multicolumn{1}{r|} {34.21}                      & 15.64                     & 17.88                     & 22.51                     \\
ASTGCN                                       & 25.09                     & 28.17                     & \multicolumn{1}{r|} {33.68}                     & 16.48                     & 18.66                     & 22.83                     \\
GMAN                                         & 22.88                     & \textbf{24.02}            & \multicolumn{1}{r|} {\textbf{25.96}}             & \textbf{13.80}            & \textbf{14.62}            & \textbf{15.72}            \\
\midrule
SpecSTG                                      & \textbf{22.23}            & {\underline {24.77}}               & \multicolumn{1}{r|} {{\underline {29.90}}}               & {\underline {14.93}}               & {\underline {16.70}}               & {\underline {20.25}}               \\ \bottomrule
\end{tabular}}
\label{tab:deter}
\end{table}

\begin{figure*}[t!]
    \centering
    \includegraphics[width=1\linewidth, height = 22cm]{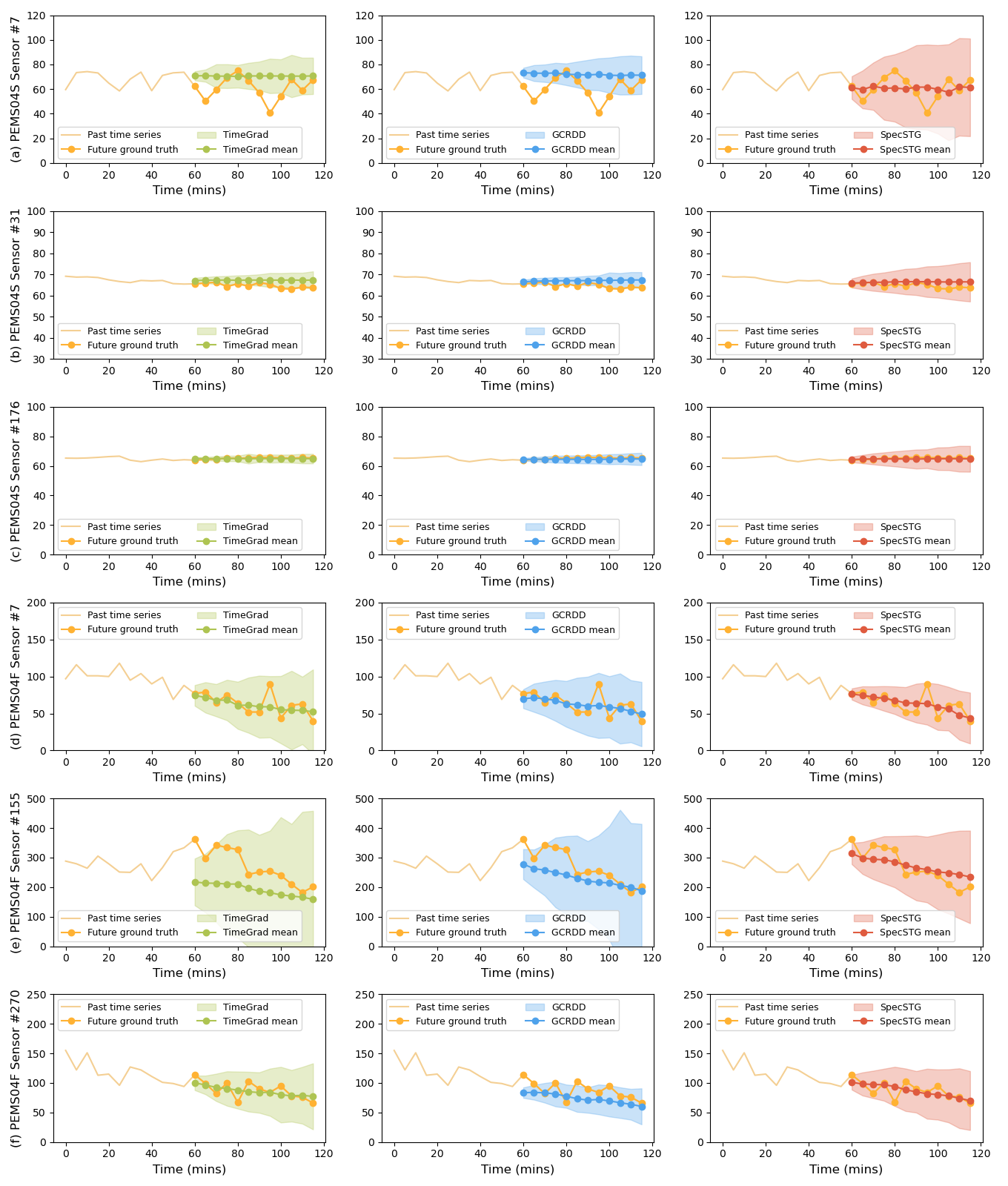}
    \caption{Supplementary forecasting visualizations on traffic speed data ((a), (b), (c)) and traffic flow data ((d), (e), (f)).}
    \label{fig:sup_visual}
\end{figure*}

\section*{Appendix F. Supplementary Forecasting Visualizations} \label{apx:visual}
In \textbf{Figure \ref{fig:sup_visual}}, we provide more forecasting visualizations of TimeGrad, GCRDD, and SpecSTG. Consistent with our conclusion in Subsection \ref{sec:visual}, SpecSTG generates more compatible forecasting intervals for data with more fluctuations (see \textbf{Figure \ref{fig:sup_visual} (a)}, \textbf{(d)}, \textbf{(e)}, and \textbf{(f)}). On the contrary, when data variations are small, the method is less effective due to the lack of systematic information in the Fourier representation (see \textbf{Figure \ref{fig:sup_visual} (b)} and \textbf{(c)}).

\section*{Appendix G. Extended Sensitivity Analysis}\label{apx:sens_analysis}
\begin{figure}[h!]
    \centering
    \includegraphics[width=1\linewidth]{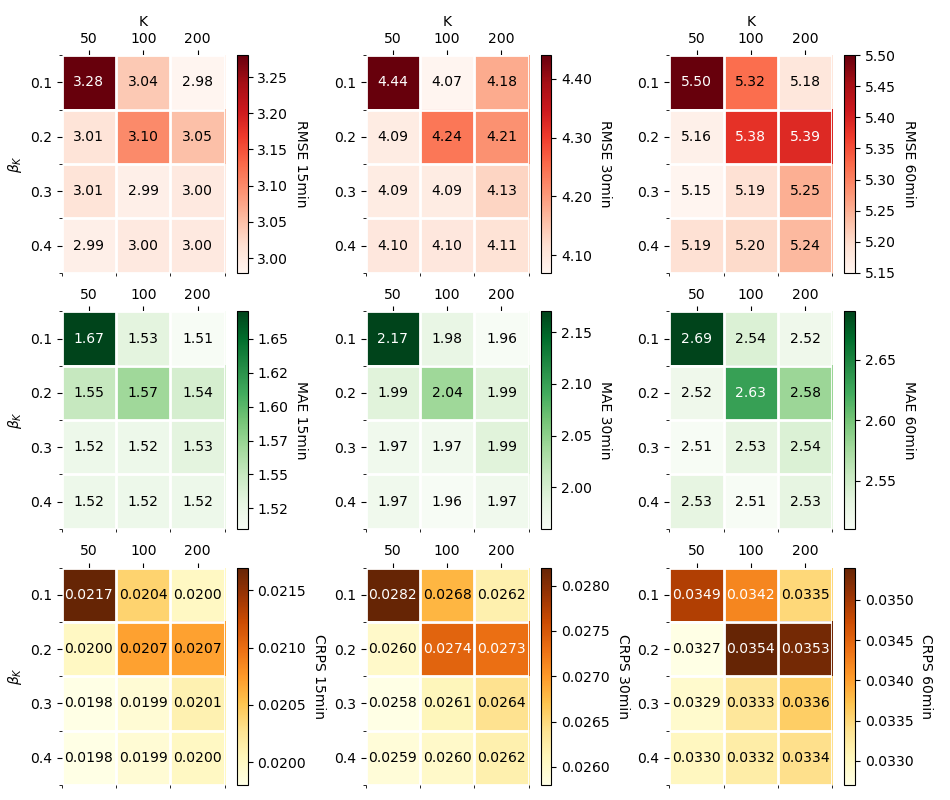}
    \caption{Extended sensitivity analysis on key hyperparameters: diffusion step $K$ and the end of beta schedule $\beta_K$ (point evaluation results at 15/30/60 minutes).}\label{fig:apx_hyper}
\end{figure}

In \textbf{Figure \ref{fig:apx_hyper}}, we show extended sensitivity analysis results for point evaluation at 15/30/60 minutes. The implementation details of this extended sensitivity analysis are consistent with those described in Subsection \ref{sec:sens_analysis}. As previously observed, SpecSTG demonstrates better performance with relatively larger values of $\beta_K$. The number of diffusion steps $K$ shows less impact on performance compared to $\beta_K$. Additionally, the forecasting outcomes exhibit minimal variation across different hyperparameter configurations. This indicates that SpecSTG is not highly sensitive to hyperparameters, thereby simplifying the hyperparameter selection process during training.

\end{document}